\documentclass{article}

\usepackage{hyperref}
\usepackage{xurl}
\usepackage[preprint]{icml2027}
\usepackage{graphicx}
\usepackage{booktabs}
\usepackage{amsmath,amssymb}
\usepackage{microtype}

\icmltitlerunning{Towards One-Loop Inference for Looped Language Models}
\definecolor{jumpaccent}{HTML}{8B5222}
\newcommand{\jump}[1]{\textcolor{jumpaccent}{#1}}

\DeclareMathOperator{\KL}{KL}
\DeclareMathOperator{\CE}{CE}
\DeclareMathOperator{\RMS}{RMS}
\newcommand{\inklingmini}{Inkling~2~Mini}
\newcommand{\inklingminiwithsize}{\mbox{\inklingmini{}-26.4B}}

\begin{document}

\twocolumn[
\icmltitle{Towards One-Loop Inference for Looped Language Models}
\begin{icmlauthorlist}
  \icmlauthor{Yifei Zhao}{iffs}
  \icmlauthor{Ying Tang}{iffs,physics}
\end{icmlauthorlist}
\icmlaffiliation{iffs}{Institute of Fundamental and Frontier Sciences, University of Electronic Science and Technology of China, Chengdu 611731, China}
\icmlaffiliation{physics}{School of Physics, University of Electronic Science and Technology of China, Chengdu 611731, China}
\icmlcorrespondingauthor{Yifei Zhao}{202521210324@std.uestc.edu.cn}
\icmlcorrespondingauthor{Ying Tang}{jamestang23@gmail.com}
\icmlkeywords{Looped Language Models, Latent Reasoning, Inference Acceleration}
\vskip 0.3in
]
\printAffiliationsAndNotice{}

\begin{abstract}
Looped language models refine predictions through repeated hidden-state updates, but serial loops make inference costly. Can the effect of many loops be recovered with far fewer? We propose a paradigm in which a lightweight predictor uses the first native loop's output to estimate the final state, or the penultimate state followed by a final native loop to improve agreement with full-loop outputs. We provide conditions for reconstructing late states and a criterion for when to use the final native loop. On Looped GPT-BERT, keeping only the first and final loops lowers WikiText-2 loss and yields around a fivefold GPU forward speedup over the original twelve loops. Across four recent architectures, at most two loops bring accuracy within three percentage points of the three-loop baseline in all 12 settings, with a \(1.34\)-\(1.50\times\) GPU forward speedup on text continuation. On a synthetic task for studying chain-of-thought performance, applying Inkling's final loop to a predicted state raises the number of correct answers from 77 to 115 out of 128. These results support using predictors with frozen pretrained looped models to reduce the number of loops with comparable or better inference.
\end{abstract}

\section{Introduction}

Looped language models repeatedly apply a shared block to refine predictions at test time \citep{dehghani2019universal,geiping2025scaling,zhu2025ouro}. This increases effective depth without adding backbone parameters. Each loop takes the previous state as input, so extra depth adds serial computation. We ask whether the first loop provides enough information to predict a later hidden state and when retaining the final loop improves the answer.

\begin{figure*}[t]
 \centering
 \includegraphics[width=\textwidth]{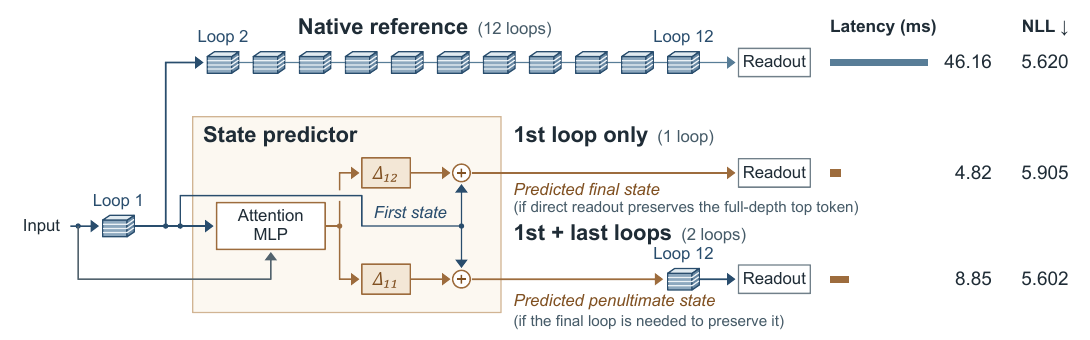}
 \caption{Framework of loop skipping demonstrated in Looped GPT-BERT. Each stack runs the shared four-layer block once. Two alternative shortcuts add an RMS-scaled residual \(\Delta_j\) to the first-loop state. \emph{1st loop only} (direct readout) predicts the final state for readout. \emph{1st + last loops} (retain final loop) predicts the loop-11 state and runs loop 12. We report median GPU forward latency (released weights, one 128-token window) and WikiText-2 test NLL (nats/token). On Inkling, the last loop improves three-hop accuracy at a fixed predicted state (Figure~\ref{fig:transfer}b), consistent with gains from additional loops on multi-hop lookup \citep{thinking2026inkling}.}
 \label{fig:architecture}
\end{figure*}

When some components of a hidden state settle quickly and the rest change slowly, estimating the slow components can be enough to reconstruct the fast components near equilibrium. We test this idea in a language model whose explicit energy measures deviation from that equilibrium. Textual chain-of-thought (CoT) passes information from earlier steps through tokens, and Coconut passes it through hidden states \citep{wei2022cot,hao2024coconut}. This motivates a test of whether a predicted state can serve as an intermediate result that the final loop can use.

Floodgates drafts future tokens in parallel with shallow loops and verifies them at full depth during decoding \citep{suleymanzade2026floodgates}. Both of our shortcuts start with the first native loop and predict a late state (Figure~\ref{fig:architecture}). \emph{1st loop only} reads out the predicted final state. \emph{1st + last loops} runs the final native loop from the predicted penultimate state. We fit the predictor to native late states and include its runtime in GPU comparisons at matched native loop counts.

The composition of native updates defines a map from an early state to a late one. We approximate this map with a single learned step and examine how the original model can use the predicted state. We make three contributions.

\begin{itemize}
\setlength{\itemsep}{2pt}
\item \textbf{Skipping intermediate loops.} Across architectures, predicting a late state from the first loop enables inference with one or two native loops. On Looped GPT-BERT, the two-loop path lowers WikiText-2 loss relative to a same-capacity learned early readout.
\item \textbf{Criterion on when the approximation suffices.} Under fast relaxation, we bound the error of reconstructing late states from their slow components. This bound, readout sensitivity and the logit margin give a sufficient condition for preserving the full model's top prediction.
\item \textbf{Predicting intermediate states for chain-of-thought.} The original model can continue reasoning from a predicted state. In \inklingmini{}, running the final loop improves three-hop accuracy over direct readout from the same state (Figure~\ref{fig:transfer}b).
\end{itemize}

Code is hosted in the \href{https://github.com/Machine-learning-and-complex-systems/Towards-One-Loop-Inference-for-Looped-Language-Models}{GitHub repository}.

\section{Related Work}

\paragraph{Shortcuts for inference.}
LayerSkip trains shallow exits and uses later layers to verify draft tokens \citep{elhoushi2024layerskip}. LoopFormer learns updates of different step sizes along the sequence of loop states \citep{jeddi2026loopformer}. LoopFold predicts later loop inputs from the first loop, then runs backbone calls in parallel and corrects their outputs \citep{song2026loopfold}. Late Entry allows new tokens to enter at a later loop using a full-depth prefix cache \citep{jin2026late}. Our GPT-BERT paths omit the middle backbone calls across the input window and reuse the original readout, with one optional final loop.

\paragraph{Energy-based models and fixed points.}
Explicit energies guide attention updates and the refinement of candidate predictions or latent reasoning states \citep{hoover2023energy,gladstone2025ebt,chen2026ebr}. CRAM searches over discrete candidates, and ERM uses Hopfield-type memories to score trajectories and guide search and depth selection \citep{hoover2026cram,zhao2026erm}. Phase kernels reshape the energy landscape of dense associative memories and improve retrieval at high memory loads \citep{zhao2026phase}. Deep Equilibrium Models define layer outputs as fixed points \citep{bai2019deq}. Equilibrium Reasoners learn attractors for reasoning, and Fixed-Point Reasoners use convergence to decide when to stop \citep{huang2026eqr,movahedi2026fpr}. In our controlled model, minimizing an explicit energy gives the equilibrium used to reconstruct the fast part of a predicted state.

\paragraph{Chain-of-thought and latent reasoning.}
Textual CoT makes intermediate results available as tokens, and Coconut feeds hidden states back as continuous inputs \citep{wei2022cot,hao2024coconut}. Looped Transformers with a scratchpad can simulate certain explicit chains of thought \citep{saunshi2025latent}. Causal abstraction and work on permutation traversal use interventions to study how intermediate states affect later computation \citep{geiger2025causal,sesodia2026inside}. On relation chains, we compare direct readout with running the final loop from the same predicted state.

\paragraph{Loops and residual connections.}
Looped language models reuse parameters across loops \citep{fan2026gptbert,fishman2026dory}. Adaptive computation can learn when to stop \citep{graves2016act}. Attention Residuals lets layers attend to earlier representations \citep{kimi2026residuals}. ARLoop retains the input embedding and recent residual blocks, and LoopRes attends to recent loops when adapting pretrained models \citep{yao2026arloop,zhao2026loopres}. Loopback studies what shared blocks store and read through residual connections \citep{ouyang2026loopback}.
Mixture-of-Loops Transformer (MoLT) selectively reuses attention keys and values across layers and loops \citep{zhu2026molt}. Our shortcuts predict late states from the first loop of an existing backbone.
Layerwise fixed points and the effects of architectural choices have also been studied \citep{blayney2026looped,zhuang2026recurrence}. We test both shortcuts across models that pass information between loops in different ways.

\section{Method}
\label{sec:method}

\subsection{Predict a Late State, Then Read or Continue}

Let \(x\) denote a fixed input and \(a_x\) its fixed representation. Let \(s_k\) denote the dynamic state needed by the next update or readout after loop \(k\), including residual history or messages passed between loops as required. A native run of depth \(R\) applies \(s_k=F_{k,x}(s_{k-1})\) and returns \(D_x(s_R)\), where \(D_x\) is the native logit readout. The index \(k\) also accommodates models with loop-dependent operations. Both shortcuts execute \(F_{1,x}\) and use a predictor \(P_j\) to estimate the state after loop \(j\) as \(\hat s_j=P_j(s_1,a_x)\). The two paths produce logits
\begin{equation}
 z_{\mathrm{jump}}=D_x(\hat s_R),\quad
 z_{\mathrm{tail}}=D_x(F_{R,x}(\hat s_{R-1})).
 \label{eq:paths}
\end{equation}
The \emph{1st loop only} path uses one native loop. The \emph{1st + last loops} path uses two and supplies the state interface expected at loop \(R\). Ground-truth late-state replay checks that this interface reproduces the reference computation.

The predictor learns a residual relative to the first native state. For transferred language checkpoints, we use an affine map for each predicted write and reuse the available prefix interface. For four-option tasks, we initialize the maps by ridge regression and tune them through the frozen readout or final-loop path. We merge the learned correction into the affine maps before inference. For Looped GPT-BERT we use
\begin{equation}
\hat s_j=s_1+\RMS(s_1)\, W_j(H(s_1,a_x)),
\label{eq:predictor}
\end{equation}
where \(\RMS(s_1)\) is the root mean square across hidden coordinates at each token. The shared body \(H\) uses masked attention and a feed-forward block, and \(W_j\) is a zero-initialized affine head. The two heads target \(j=11,12\) and only the selected head runs during inference. The model uses the same four-layer block in every native loop. Retaining the final loop applies that same block to the predicted penultimate state.

We train both GPT-BERT paths on native late states, teacher outputs, and observed tokens. The final-loop loss backpropagates through the last native update to the predictor. We fit the predictor on frozen released weights or alongside the backbone during the latter half of training from scratch. In the latter case, we detach the teacher states and update the backbone using its own twelve-loop objective. Appendix~\ref{app:protocol} gives the objectives and schedules.

\subsection{Fast Relaxation Makes Late States Predictable}
\label{sec:theory}

At fixed \(x\), suppose the local state splits into slowly varying \(u_k\) and faster \(v_k\). On the relevant segment assume
\(v_{k+1}=Av_k+Bu_k+c+r_k\), where \(A\) is the fast-state transition, \(B\) couples the slow state to the fast update, \(c\) is an offset, and \(r_k\) is a residual. Assume \(\|A\|\le q<1\) in the induced operator norm and \(\|r_k\|\le\eta\). Holding \(u\) fixed gives the affine equilibrium \(v_*(u)=(I-A)^{-1}(Bu+c)\), with identity matrix \(I\). If \(v_*\) is \(L\)-Lipschitz and \(\|u_{k+1}-u_k\|\le\epsilon\), the deviation \(e_k=v_k-v_*(u_k)\) satisfies \(\|e_{k+1}\|\le q\|e_k\|+L\epsilon+\eta\). Iterating yields
\begin{equation}
 \|e_R\|\le q^{R-1}\|e_1\|+
 (L\epsilon+\eta)\frac{1-q^{R-1}}{1-q}.
 \label{eq:fast}
\end{equation}
The first term decays with depth, and the second is controlled by slow drift and local approximation error. Given a prediction \(\hat u_R\) of \(u_R\), reconstructing the fast state as \(v_*(\hat u_R)\) incurs an error of at most \(\|e_R\|+L\|\hat u_R-u_R\|\). The contraction condition applies to the fast component.

\subsection{Readout Sensitivity Determines the Cost of Error}

For direct readout, let \(G=D_x\) and \(s=s_R\). For final-loop continuation, let \(G=D_x\circ F_{R,x}\) and \(s=s_{R-1}\). The corresponding predicted state is \(\hat s\). With reference top token \(a\) and competitor tokens \(b\), the minimum logit margin is \(m=\min_{b\ne a}(G_a(s)-G_b(s))>0\). Assume the margins are absolutely continuous along the segment from \(s\) to \(\hat s\) and let \(K_G\) bound the dual norm of each margin gradient along it. Integrating along the segment gives the sufficient condition
\begin{equation}
 K_G\|\hat s-s\| < m
 \label{eq:margin}
\end{equation}
for preserving the top token in that context. The bound accounts for both the prediction error and the sensitivity of the path from state to logits. A path satisfying the bound preserves the full model's top prediction. Residual decay and reuse of earlier writes motivate testing direct readout, and dependence on the last write motivates testing the final loop. We compare each shortcut with the native prefix that uses the same number of loops.

Explicit energy gives a concrete instance of fast relaxation. For a Euclidean gradient step \(s^+=s-\alpha\nabla E_x(s)\), near a smooth minimum, an error along a Hessian eigenvector of curvature \(\lambda_i\) is multiplied to first order by \(1-\alpha\lambda_i\). In a four-loop language model we set \(E(v\mid a_x,u)=\tfrac12\|v-g_\theta(a_x,u)\|_2^2\). With the updated slow state held fixed, a step of size \(0.75\) reduces fast-state distance by a factor of four and energy by a factor of sixteen. Appendix~\ref{app:theory} separates this conditional energy from local stability and from a quadratic approximation to readout divergence.

\section{Experiments}
\label{sec:exp}

\subsection{Language Models and the Final Loop}

\begin{figure*}[t]
 \centering
 \includegraphics[width=\textwidth]{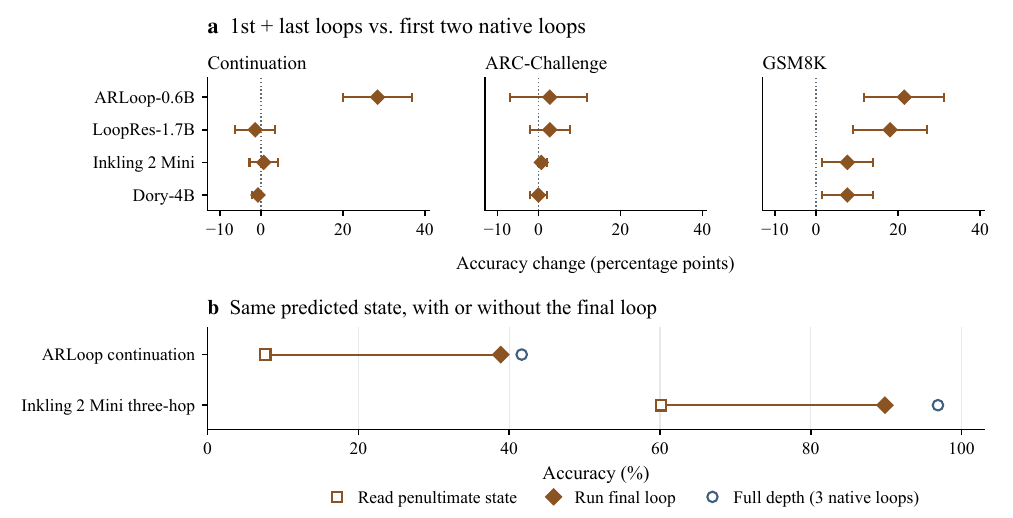}
 \caption{Language-model accuracy and the effect of the final loop. (a) Accuracy change for \emph{1st + last loops} relative to running the first two native loops. Points show paired accuracy differences with 95\% example-bootstrap intervals (\(n=144\) per setting). Continuation uses ridge fits and next-token accuracy. ARC-Challenge and GSM8K use output-tuned fits and four-option accuracy from Table~\ref{tab:four_models}. (b) Both routes start from the same predicted penultimate state. One reads it out directly and the other runs the final loop before readout. The full-depth reference runs all three native loops in both evaluations (\(n=144\) for ARLoop continuation and \(n=128\) for three-hop reasoning with \inklingminiwithsize{}).}
 \label{fig:transfer}
\end{figure*}

We evaluate text continuation, the Challenge Set of the AI2 Reasoning Challenge (ARC-Challenge) \citep{clark2018arc} and GSM8K. For each checkpoint and task, we fit an affine predictor on 72 training examples and evaluate on a separate set of 144. On the four-option tasks, we tune the maps using correct answers and the native three-loop output distribution. The residual scales and update gains are fixed from validation of the corresponding 800-question fits (Appendix~\ref{app:transfer}). We evaluate released ARLoop, LoopRes and Dory-4B checkpoints \citep{yao2026arloop,zhao2026loopres,fishman2026dory} and the Inkling~2~Mini research counterpart \citep{thinking2026inkling}, a variable-loop checkpoint we call \inklingmini{} throughout. Compared with two native loops, the shortcut scores higher in nine of the 12 settings and ties in one. The two decreases, both on continuation, are at most 1.39 percentage points (pp) (Figure~\ref{fig:transfer}a). Compared with three native loops, the shortcut matches or exceeds accuracy in seven settings and comes within 2.78 pp in the other five (Table~\ref{tab:cross}).

\begin{table*}[!t]
\centering
\caption{Four-option accuracy (\%) with affine predictors fitted on 72 questions per checkpoint and task and tuned with an output loss. Each comparison uses the same 144 test questions. ARLoop scores each candidate text by character-normalized log-likelihood. LoopRes, \inklingminiwithsize{} and Dory score answer letters. The native baselines read out after one, two or three loops.}
\label{tab:four_models}
\small
\setlength{\tabcolsep}{4.6pt}
\begin{tabular}{@{}llrrrrr@{}}
\toprule
& & \multicolumn{2}{c}{One native loop} & \multicolumn{2}{c}{Two native loops} & \multicolumn{1}{c}{Three native loops} \\
\cmidrule(lr){3-4}\cmidrule(lr){5-6}\cmidrule(l){7-7}
Model & Task & Native & \phantom{Direct readout (ours)}\llap{\jump{1st loop only}} & Native & \phantom{Retain final loop (ours)}\llap{\jump{1st + last loops}} & Native \\
\midrule
ARLoop-0.6B & ARC-Challenge & 25.0 & \jump{27.1} & 30.6 & \jump{33.3} & 36.1 \\
 & GSM8K & 31.3 & \jump{41.7} & 30.6 & \jump{52.1} & 44.4 \\
\addlinespace[1.5pt]
LoopRes-1.7B & ARC-Challenge & 59.0 & \jump{70.1} & 75.0 & \jump{77.8} & 75.0 \\
 & GSM8K & 23.6 & \jump{42.4} & 31.9 & \jump{50.0} & 33.3 \\
\addlinespace[1.5pt]
\mbox{Inkling-26.4B} & ARC-Challenge & 92.4 & \jump{91.7} & 96.5 & \jump{97.2} & 97.2 \\
 & GSM8K & 47.2 & \jump{45.8} & 45.8 & \jump{53.5} & 48.6 \\
\addlinespace[1.5pt]
Dory-4B & ARC-Challenge & 80.6 & \jump{79.2} & 84.0 & \jump{84.0} & 84.0 \\
 & GSM8K & 41.0 & \jump{43.1} & 46.5 & \jump{54.2} & 48.6 \\
\bottomrule
\end{tabular}
\end{table*}

\begin{figure*}[t]
 \centering
 \includegraphics[width=\textwidth]{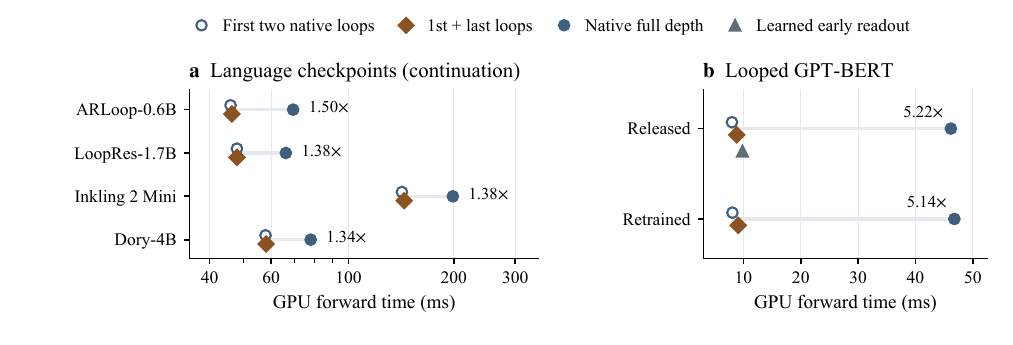}
 \caption{Measured GPU forward time. (a) Matched continuation inputs at batch size one with key--value caching disabled. The full reference uses three native loops for each checkpoint. Points show the median over seven timed inputs per route after two warm-ups. (b) Looped GPT-BERT on pretokenized 128-token windows at batch one, using the median of twenty timings. Learned early readout uses the same predictor capacity and training budget on released weights.}
 \label{fig:latency}
\end{figure*}

\paragraph{Effect of the final loop.}
We compare direct readout and final-loop continuation from the same predicted penultimate state (Figure~\ref{fig:transfer}b). On ARLoop continuation, the final loop raises the number of correct predictions from \(11/144\) to \(56/144\), compared with \(15/144\) for the first two native loops and \(60/144\) for all three. With the same final update, using the learned state in place of a copy of the first-loop state raises the count from 36 to 56. On synthetic three-hop tasks with \inklingmini{}, the final loop raises the count from \(77/128\) to \(115/128\), with \(124/128\) at three native loops. Thinking Machines Lab found that additional loops help the same research model follow longer lookup chains \citep{thinking2026inkling}. Appendix~\ref{app:cot} reports the depth sweep and tests whether replacing the predicted prefix state changes the answer to the downstream relation chain.

\paragraph{One versus two loops.}
Table~\ref{tab:four_models} compares both shortcuts with native runs at the same loop count. \emph{1st loop only} scores higher than one native loop in five of eight settings. \emph{1st + last loops} scores higher than two native loops in seven settings and ties on Dory ARC-Challenge. On GSM8K, the shortcut exceeds three native loops by 4.86--16.67 pp across the four checkpoints. Its agreement with three native loops ranges from 47.9\% to 100\% across the eight four-option settings (Table~\ref{tab:cross}). In Appendix~\ref{app:transfer}, Table~\ref{tab:four_models_ridge} reports results on the same questions with predictors fitted solely to native late-loop states.

In a separate intervention on ARLoop, LoopRes and \inklingmini{}, restoring the last native write increases agreement by 11.1--45.1 pp across the six settings. Restoring the previous write changes agreement by \(-1.4\) to \(+2.1\) pp. The larger effect of restoring the last write points to the role of the final update in producing the readout (Table~\ref{tab:four_diagnostics} and Figure~\ref{fig:diagnostics}a).

On text continuation, \emph{1st + last loops} gives a \(1.34\)--\(1.50\times\) forward speedup over three native loops across the four checkpoints (Figure~\ref{fig:latency}a and Table~\ref{tab:transfer_time}). At the same two-loop budget, prediction adds at most 2.10 ms to the measured forward time. The largest absolute saving is on \inklingmini{}, from 198.84 to 144.26 ms. On a separate 24-question test with \inklingmini{}, it also reduces free-response generation time from 367.5 to 269.4 ms per response (Appendix~\ref{app:transfer}).

\subsection{Explicit Energy and Puzzle Solvers}

\begin{figure*}[t]
 \centering
 \includegraphics[width=\textwidth]{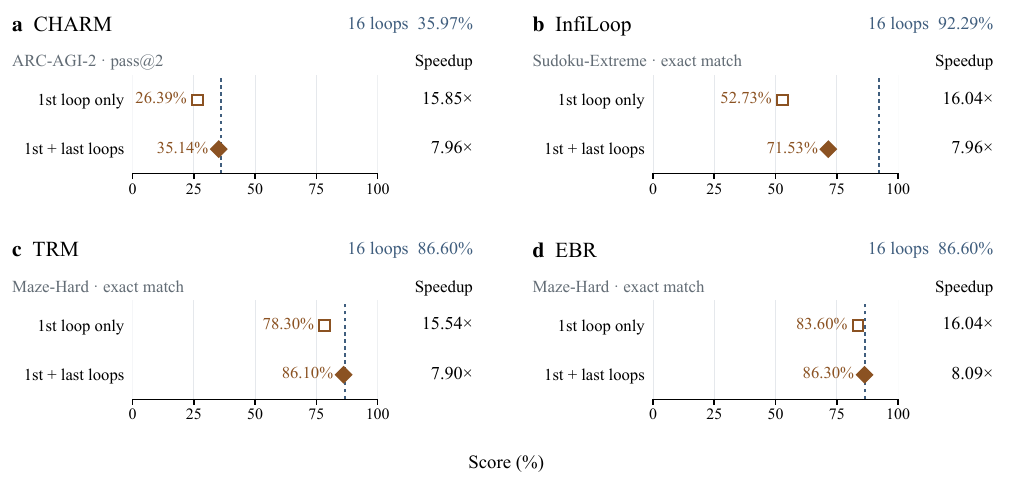}
 \caption{Puzzle scores and speedups across three tasks. Speedups are relative to each model's 16-loop reference. Dashed lines mark the reference scores. \emph{1st loop only} reads out the predicted state. \emph{1st + last loops} runs the final native loop before readout. CHARM reports task-averaged pass@2 on 120 tasks with 16 views per grid. InfiLoop reports whole-grid exact match on 2,048 Sudoku-Extreme puzzles. TRM and EBR report labeled-path exact match on the same 1,000 mazes.}
 \label{fig:solvers}
\end{figure*}

In a four-loop language model with explicit conditional energy, reconstructing the fast component at the same predicted slow state increases full-depth agreement from 81.57\% to 83.01\% for direct readout and from 89.18\% to 89.61\% with the final loop. The fast update reduces conditional energy by a factor of sixteen. We test whether the shortcuts retain task performance at lower latency across three tasks and four solvers, using each model's 16-loop reference (Figure~\ref{fig:solvers}). To compare how different final updates change state error, we use energy-based and fixed-point solvers on the same Sudoku-Extreme puzzles (Figure~\ref{fig:diagnostics}c).

The two-loop shortcut reaches 35.14\% pass@2 with CHARM on the second version of the Abstraction and Reasoning Corpus for Artificial General Intelligence (ARC-AGI-2), compared with 35.97\% at full depth (Figure~\ref{fig:solvers}). On Maze-Hard it reaches 86.10\% for TRM and 86.30\% for EBR, compared with 86.60\% for both at full depth. The three models give speedups of \(7.90\)--\(8.09\times\) over 16 native loops (Table~\ref{tab:solver_routes}).

On Sudoku-Extreme, applying InfiLoop's final loop to the predicted state raises whole-grid accuracy from 52.73\% to 71.53\%. The two-loop shortcut gives a \(7.96\times\) speedup over 16 loops. The native two- and 16-loop scores are 72.27\% and 92.29\%, respectively (Figure~\ref{fig:solvers}b and Table~\ref{tab:sudoku_routes}).

\begin{figure*}[t]
 \centering
 \includegraphics[width=\textwidth]{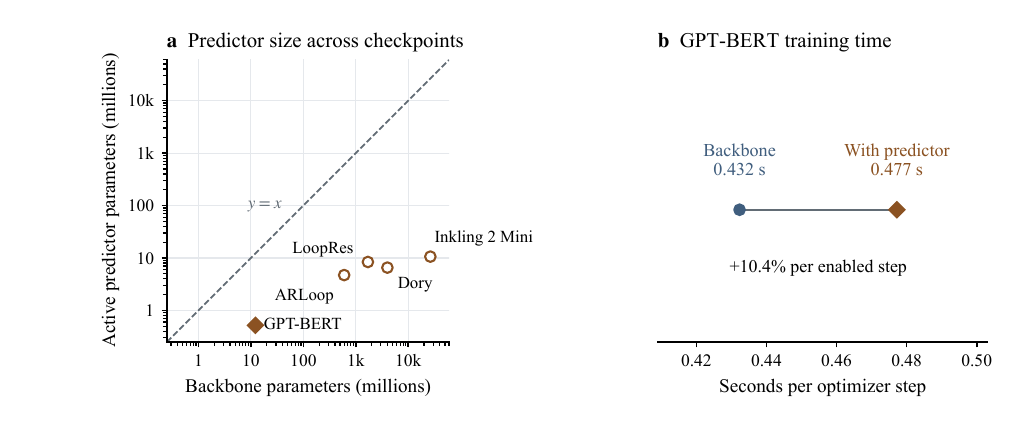}
 \caption{Predictor size and training cost. (a) Parameters used by \emph{1st + last loops}. The filled diamond denotes GPT-BERT's attention predictor with one active output head. Open circles denote the affine residual maps used at inference. Backbone sizes are total parameters. The dashed \(y=x\) line marks equal predictor and backbone size. (b) Median time per optimizer step in 32-step intervals in the same GPT-BERT training run, before and after enabling the predictor.}
 \label{fig:cost}
\end{figure*}

\subsection{Inference Time and Predictor Cost}

\paragraph{A matched training comparison.}
Looped GPT-BERT has 12.18M backbone parameters and runs twelve loops \citep{fan2026gptbert}. We fit a 594,048-parameter predictor on released weights and repeat the experiment with a backbone trained from scratch. Relative to two native loops, the shortcut lowers WikiText-2 negative log-likelihood (NLL) by 0.437 and 0.433 nat/token and raises accuracy on the Benchmark of Linguistic Minimal Pairs (BLiMP) by 2.18 and 3.88 pp, respectively. Forward times are 8.85 versus 46.16 ms on the released model and 9.11 versus 46.78 ms on the retrained model, giving \(5.22\times\) and \(5.14\times\) speedups over twelve loops. The same-capacity learned early readout on released weights takes 9.83 ms and has 0.088 higher NLL than retaining the final loop. Appendix~\ref{app:protocol} reports the complete depth sweep and the learned readout comparison.

\paragraph{Training overhead and scale.}
The predictor is enabled for 2,120 of the 4,220 backbone training steps. Median step time rises from 0.432 to 0.477 s between the two phases (Figure~\ref{fig:cost}b). The released-weight predictor takes 56.4 s to fit, including generation of the targets. At batch one, the time saved over twelve native loops adds up to this fitting time after about 1,510 windows. On the four larger frozen models we fit one affine map per predicted residual. For continuation we sample 4,096 token positions per checkpoint. Each map at hidden width \(h\) has \(h(h+1)\) coefficients. The four larger checkpoints use 4.72M to 10.62M coefficients per task, depending on width and the number of predicted residuals (Figure~\ref{fig:cost}a). The neural predictor's training time is measured on GPT-BERT.

\section{Discussion}

At the same loop count, accuracy depends on where the predicted state enters the model. In GPT-BERT, feeding the predicted state into the final loop lowers WikiText-2 loss relative to a same-capacity learned early readout. In \inklingmini{}, the final loop answers more three-hop questions correctly from the same predicted state. In the four-loop energy model, reconstructing the fast component at a fixed predicted slow state lowers language modeling loss. The Sudoku probes show that state error and answer accuracy can respond differently to a final update.

Output distillation trains a student on a teacher's predictions \citep{hinton2015distilling}. In our output-tuned fits, the teacher distribution supervises a map between loop states, and the frozen backbone reads out the predicted state or continues from it.

Validation guided the choice of prediction objective and architecture in GPT-BERT. The initial screen trained each predictor for 256 steps on packed 128-token windows. With the final loop, adding output losses and changing the state-loss weight reduced validation NLL from 6.0791 to 5.5677. With output losses, the attention predictor reached 5.5393. A separate WikiText-2 validation comparison selected the twelfth-loop teacher for the released-weight test (Figure~\ref{fig:development}).

Predicted states can initialize other inference procedures. For fixed-point and energy solvers, the effect of this initialization depends on the subsequent update and on the readout (Appendix~\ref{app:energy}). A phase-kernel associative memory could use a predicted state to initialize retrieval \citep{zhao2026phase}. For speculative decoding, a state predictor could draft tokens for full-depth verification, as used in Floodgates \citep{suleymanzade2026floodgates}. The Dory-4B ablation shows that deeper prefix keys and values can improve one-loop agreement (Appendix~\ref{app:transfer}).

Our evaluation measures accuracy and runtime for fixed checkpoints and input formats. The GPT-BERT reconstruction uses one seed and follows public materials for data packing and masking. We measure forward speed at a fixed context length and report predictor training cost separately. The synthetic intervention replaces the whole prefix state to measure its effect on continuation along the relation chain. Future evaluations can use longer cached decoding and test route selection on held-out architectures.

\section{Conclusion}

A learned map from the first-loop state to a late state supports inference with one or two native loops. At matched loop counts, retaining the final loop lowers GPT-BERT loss and improves accuracy in seven of eight four-option settings. It also improves ARLoop continuation and free-response GSM8K with \inklingmini{}. Fast relaxation gives conditions for reconstructing late states, and the readout bound relates state error to preservation of the top prediction. The choice of retained loops can be evaluated against both accuracy and forward time.

\section*{Impact Statement}
Predicting late states can reduce forward latency by skipping intermediate loops. Accuracy can change in either direction, depending on the task and checkpoint. Deployment evaluations can measure both effects on their own inputs and hardware.

\section*{AI Disclosure Statement}
Generative AI assisted with experimental design and controls, implementation and debugging, mathematical exposition and interpretation, literature synthesis, and drafting the manuscript and figures. The measurements come from archived runs of the evaluated models. Numerical claims and plots were checked against the run records, and citations against the primary sources. AI-assisted reviews identified claims for further checking and informed revisions. The authors are responsible for the final text, mathematical claims, code, and results.

\bibliography{references}
\bibliographystyle{icml2027}

\clearpage
\appendix
\section{Notation}
\label{app:notation}

\noindent\begin{minipage}{\columnwidth}
\centering
\captionof{table}{Notation used in the analysis.}
\label{tab:notation}
\begin{tabular}{@{}lp{0.71\columnwidth}@{}}
\toprule
Symbol & Definition \\
\midrule
\(x,a_x\) & Input and fixed input representation \\
\(k,j,R\) & Loop index, prediction target index, and native depth \\
\(s_k,\hat s_j,\tilde s_R\) & Native state, predicted late state, and final state recomputed from the predicted input \\
\(F_{k,x},D_x,G\) & Native update, native readout, and path from an interface state to logits \\
\(P_j,H,W_j\) & Late-state predictor, shared predictor body, and residual head \\
\(z_t,p_k,\hat p_t,y\) & Logits of path \(t\), native loop-\(k\) distribution, shortcut distribution, and observed token \\
\(\mathcal L_s,d,h\) & State loss, state coordinates per token, and hidden width of an affine residual map \\
\(\Delta_j,\RMS\) & Predicted residual at loop \(j\) and root mean square \\
\(\CE,\KL\) & Cross-entropy and Kullback--Leibler divergence \\
\(u_k,v_k,v_*(u)\) & Slow state, fast state, and conditional fast equilibrium \\
\(A,B,c,r_k\) & Fast transition, coupling, offset, and local residual \\
\(I,\hat u_R\) & Identity matrix and predicted slow state at depth \(R\) \\
\(q,L,\epsilon,\eta\) & Fast gain bound, equilibrium Lipschitz constant, slow drift bound, and residual bound \\
\(e_k,\delta\) & Deviation from fast equilibrium and interface prediction error \\
\(a,b,m,K_G\) & Reference top token, competitor, minimum logit margin, and bound on margin-gradient dual norm \\
\(E_x,E,\alpha,\lambda_i\) & Energy, conditional fast energy, gradient step length, and Hessian curvature \\
\(g_\theta\) & Learned map from input and slow state to the fast equilibrium \\
\(J,p,p_\tau,\Delta,\xi\) & Readout Jacobian, reference and interpolated distributions, logit change, and perturbation amplitude \\
\(C_p\) & Categorical covariance at probability vector \(p\) \\
\(\Lambda,\phi\) & Linear skip and nonlinear part of a fast-state update \\
\(s_*,A_*,Q,M\) & Fixed point, its linearization, positive definite weighting, and Lyapunov matrix \\
\(H_{ab}\) & Hessian block of the conditional energy with variables \(a,b\in\{u,v\}\) \\
\bottomrule
\end{tabular}
\end{minipage}
\par\medskip

\section{GPT-BERT Training and Evaluation}
\label{app:protocol}

\paragraph{Model and data.}
The public GPT-BERT backbone contains 12,181,780 parameters, a width of 384, six attention heads per physical layer, an 8,192-token vocabulary and 128-token contexts \citep{fan2026gptbert}. We use the \href{https://huggingface.co/Survivor613/BabyLM2026-Strict-Small-looped_GPT-BERT/tree/422904b427051afea7da545f8f74a00178c89c50}{authors' checkpoint at revision \texttt{422904b}} and preserve its tied readout. The fixed input is added at every loop, and dynamic weighted accumulation mixes signals inside each physical layer. A 192-dimensional, four-head masked attention predictor receives normalized \(s_1\) and \(a_x\), augmented with their log RMS values and cosine similarity. Its feed-forward block has width 384. We predict residuals relative to \(s_1\) with two zero-initialized affine heads. The wrapped path matches the released model's logits exactly on the replay audit, and \(F_{12,x}(s_{11})\) reproduces the native twelfth state. Applying the shared update to a copy of \(s_1\) gives native two-loop output.

The model card specifies 4,220 updates, 256 sequences per update and ten epochs. The \href{https://github.com/JT-Ushio/babylm26-nlp-spring}{public cleaned source}, \texttt{new\_data2}, matches the card's 7,482,193 words under Unicode counting and 352,096 nonempty lines. The manuscript's per-source whitespace counts sum to 7,482,189 and it uses \texttt{new\_data1} as the directory label. The final training configuration file was unavailable. We reconstruct packing, mask sampling and the learning-rate schedule from the public materials. We initialize with seed 42, accumulate two 128-sequence microbatches per update, use a 1:3 ratio of masked to causal sequences, and optimize the backbone with LAMB. The peak and final learning rates are 0.0141 and 0.00141, with 41 warm-up steps, a cosine decay horizon of 2,600 steps, and a constant floor thereafter. The nominal input position count includes padding. The predictor starts at update 2,101 and runs for 2,120 updates on the same training batches. Full-trajectory teacher states are detached in the predictor optimization. The native training path uses dropout, while native continuation and readout in the predictor path run with dropout disabled.

\paragraph{Objectives.}
We average the squared state error at each valid token over \(d\) coordinates and divide by the native target's mean squared magnitude with a floor of \(0.01\). Summing over the terminal and penultimate prediction heads yields \(\mathcal L_s\). Let \(t\in\{\mathrm{jump},\mathrm{tail}\}\) index paths and \(\hat p_t\) denote the softmax of \(z_t\). The released-weight objective is
\begin{equation}
\tfrac12\mathcal L_s+\sum_t
\left[\KL(p_{12}\Vert \hat p_t)+0.25\,\CE(y,\hat p_t)\right].
\end{equation}
The reconstructed protocol uses
\begin{equation}
0.1\mathcal L_s+\sum_t
\left[\CE((p_6+p_{12})/2,\hat p_t)+0.75\,\CE(y,\hat p_t)\right].
\end{equation}
For a soft target, \(\CE\) denotes cross-entropy with that target distribution. At each update the output terms use one cyclic position-index class modulo four intersected with the valid labels. The released-weight predictors use 768 steps with microbatch size 32 and revisit batches from the first epoch, for 3,145,728 nominal positions. AdamW uses initial learning rate 0.0012, \((0.9,0.95)\) moment coefficients, and 0.01 weight decay. The single-card reconstruction uses the same predictor learning rate and moment coefficients. The learned early readout uses the second true state as input, the same predictor capacity and data, and a learned readout without a subsequent native loop.

\begin{figure*}[t]
\centering
\includegraphics[width=\textwidth]{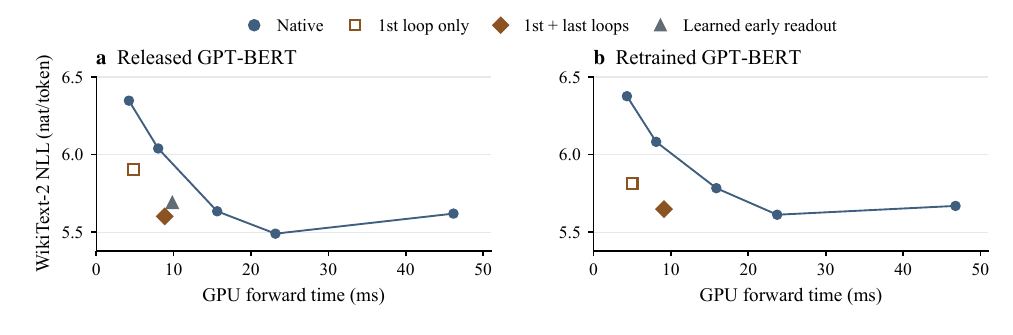}
\caption{WikiText-2 loss and GPU forward time for GPT-BERT. Native points from left to right use 1, 2, 4, 6 and 12 loops. The learned early readout receives the true second-loop state and is fitted on released weights.}
\label{fig:gpt_sweep}
\end{figure*}

\paragraph{Selection and timing.}
The released-weight predictor reached update 768 in 56.36 seconds, including teacher computation. For the retraining timing comparison, we divide elapsed time by 32 in intervals entirely before or after predictor activation, excluding startup. The medians are 0.4323 and 0.4772 seconds per update. These intervals come from one run with the same global batch size. At inference, the shortcut that retains the final loop uses 519,936 predictor parameters, including one output head. The training predictor has 594,048 parameters across both output heads.

An initial 256-step development screen compares tokenwise state regression, tokenwise prediction with output losses, and the attention predictor with output losses. We then compare the final-loop designs on 192 WikiText-2 validation windows. Native two loops, learned early readout, the attention predictor with the twelfth-loop teacher, and the attention predictor with a mixed sixth- and twelfth-loop teacher give NLLs of 5.84734, 5.50818, 5.39666, and 5.41141, respectively. The locked released-weight design uses the twelfth-loop teacher. The primary criterion set for that test was an upper NLL-difference confidence bound below zero against native two loops, at most a one-point BLiMP accuracy loss against two loops, and at least a twofold forward speedup against native twelve loops. The 0.05 nat/token tolerance against native twelve loops refers only to WikiText-2 loss.

\begin{figure*}[t]
\centering
\includegraphics[width=\textwidth]{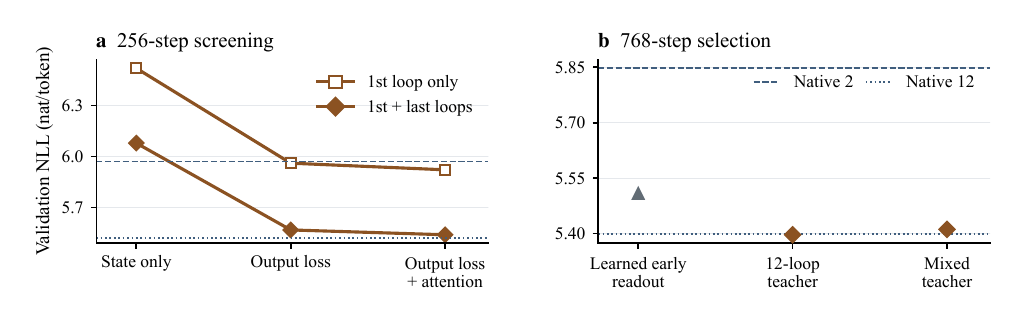}
\caption{GPT-BERT validation before test evaluation. (a) Initial predictor screen on 64 windows. The output-loss configuration also changes the state-loss weight. (b) Extended selection on 192 windows. Both teacher configurations use \emph{1st + last loops}. Blue lines denote native two and twelve loops in each protocol.}
\label{fig:development}
\end{figure*}

The WikiText-2 test set \citep{merity2017wikitext} yields 2,848 consecutive 128-token windows and 364,544 target positions. Each nonempty source row contributes a beginning- and end-of-sequence token to the stream, and these positions count in the score. We resample contiguous blocks of eight windows 5,000 times for paired NLL intervals. The BabyLM-filtered version of BLiMP \citep{warstadt2020blimp} uses 59,875 minimal pairs over 67 categories, with a comparison of summed causal sentence scores and half credit for ties. Across inference paths, single-window forward timing uses pretokenized inputs, BF16 autocast and CUDA events, with four warm-ups and the median of twenty measurements. A separate batch-32 test uses 20 warm-ups and 100 measurements on the released weights. Native two loops, retaining the final loop and native twelve loops take 8.998, 9.860 and 51.805 ms, respectively.

\begin{table*}[t]
\centering
\caption{GPT-BERT quality at matched native loop counts and at full depth. WikiText-2 NLL uses nat per token. BLiMP is macro-averaged across 67 categories. The learned early readout has the same predictor capacity and training updates as the released-weight shortcut.}
\label{tab:main}
\setlength{\tabcolsep}{6pt}
\begin{tabular}{l r rr rr}
\toprule
& & \multicolumn{2}{c}{Released weights} & \multicolumn{2}{c}{Reconstructed training}\\
\cmidrule(lr){3-4}\cmidrule(lr){5-6}
Inference path & Loops & NLL $\downarrow$ & BLiMP $\uparrow$ & NLL $\downarrow$ & BLiMP $\uparrow$ \\
\midrule
Native one loop & 1 & 6.347 & 62.48 & 6.376 & 61.80\\
\jump{1st loop only} & 1 & \jump{5.905} & \jump{62.40} & \jump{5.815} & \jump{64.63}\\
Native two loops & 2 & 6.039 & 65.21 & 6.082 & 63.74\\
Learned early readout & 2 & 5.691 & 65.44 & -- & --\\
\jump{1st + last loops} & 2 & \jump{5.602} & \jump{67.38} & \jump{5.649} & \jump{67.62}\\
Native six loops & 6 & 5.491 & -- & 5.613 & --\\
Native twelve loops & 12 & 5.620 & 71.19 & 5.670 & 70.37\\
\bottomrule
\end{tabular}
\end{table*}

\begin{table*}[t]
\centering
\caption{Complete WikiText-2 depth and latency sweep. Each cell reports NLL / next-token accuracy (\%) / agreement with native twelve-loop output (\%) / median GPU forward latency (ms).}
\label{tab:fullgpt}
\small
\begin{tabular}{l r cc}
\toprule
Path & Native loops & Released weights & Reconstructed training \\
\midrule
Native one loop & 1 & 6.3473 / 14.62 / 29.18 / 4.227 & 6.3760 / 14.78 / 30.62 / 4.318 \\
\jump{1st loop only} & 1 & 5.9051 / 15.48 / 35.63 / 4.825 & 5.8155 / 16.97 / 37.87 / 5.000 \\
Native two loops & 2 & 6.0395 / 17.56 / 36.37 / 8.010 & 6.0821 / 17.65 / 36.09 / 8.092 \\
Learned early readout & 2 & 5.6905 / 18.04 / 42.41 / 9.833 & -- \\
\jump{1st + last loops} & 2 & 5.6022 / 20.01 / 53.48 / 8.847 & 5.6488 / 20.03 / 52.71 / 9.109 \\
Native four loops & 4 & 5.6349 / 21.18 / 48.30 / 15.633 & 5.7838 / 20.32 / 44.01 / 15.874 \\
Native six loops & 6 & 5.4908 / 22.26 / 58.46 / 23.160 & 5.6127 / 21.69 / 53.21 / 23.734 \\
Native twelve loops & 12 & 5.6199 / 20.32 / 100 / 46.164 & 5.6695 / 20.58 / 100 / 46.778 \\
\bottomrule
\end{tabular}
\end{table*}

\section{Additional Transfer Tests and Interventions}
\label{app:transfer}

\paragraph{Backbone provenance.}
Table~\ref{tab:checkpoints} identifies the weights used in the language experiments. We freeze the external backbones and fit the shortcut separately. For backbones we train, all inference paths within a comparison share the same resulting weights. The reconstructed GPT-BERT also trains its predictor with detached native states as described in Appendix~\ref{app:protocol}. We list fixed revisions where the run records identify them.

\begin{table*}[t]
\centering
\caption{Backbones for the language experiments. From-scratch weights were trained in this study.}
\label{tab:checkpoints}
\small
\begin{tabular}{@{}p{0.24\textwidth}p{0.19\textwidth}p{0.52\textwidth}@{}}
\toprule
Backbone & Weights & Source and configuration \\
\midrule
Looped GPT-BERT & Public, frozen & \href{https://huggingface.co/Survivor613/BabyLM2026-Strict-Small-looped_GPT-BERT/tree/422904b427051afea7da545f8f74a00178c89c50}{Authors' 4-layer, 12-loop release}, revision \texttt{422904b}. \\
Looped GPT-BERT & From scratch & Same public architecture and tokenizer. One 4,220-step reconstruction on the released BabyLM data. \\
ARLoop-0.6B & Public, frozen & \href{https://huggingface.co/osunlp/ARLoop-0.6B}{Authors' 0.6B release}, evaluated at up to three loops. \\
LoopRes-Qwen3-1.7B & Public, frozen & \href{https://huggingface.co/oahzxl/LoopRes-Qwen3-1.7B}{Authors' checkpoint adapted from Qwen3-1.7B}, evaluated at up to three loops. \\
\inklingmini{} & Public, frozen & \href{https://huggingface.co/thinkingmachines/TML-d24-varloop-research/tree/c90921bd5f78df6c8168173b9f8a3a850320c594}{Thinking Machines Lab's variable-loop research checkpoint}, revision \texttt{c90921b}. \\
Dory-4B & Public, frozen & \href{https://huggingface.co/nvidia/NVIDIA-Nemotron-Labs-Dory1.0-4B-BF16}{NVIDIA's Dory1.0-4B BF16 release}. \\
Four-loop energy LM & From scratch & Two matched 5.62M-parameter training runs on the released BabyLM data, with copied and learned shortcut inputs. \\
\bottomrule
\end{tabular}
\end{table*}

\paragraph{Task formats.}
We evaluate each checkpoint on 144 held-out examples per task. The train/test sampling seeds are 31811/31821 for WikiText-2, 31911/31921 for ARC-Challenge and 32011/32021 for GSM8K. For WikiText-2, we use up to 65 tokens per eligible line, with the last token as the target. We shuffle the choices in ARC-Challenge \citep{clark2018arc} and add three distractors from seeded numerical offsets to each GSM8K answer \citep{cobbe2021gsm}. ARLoop ranks the candidates by character-normalized continuation log-likelihood. The other models rank their next-token scores for the four answer letters. For ARLoop, direct letter scoring gave B on all 144 questions at native depth two in each task, including after cyclic rotation of the options.

\paragraph{Output-tuned affine predictors.}
Table~\ref{tab:four_models} uses 72 fitting questions per checkpoint and task, drawn from the official training splits. These are the first 72 questions from the ordered training pools used for the 800-question fits. GSM8K constructs the distractors and option order after drawing the training pool with a requested size of 1,200. We initialize each affine map by ridge regression using the position that predicts the first answer token and sampling other positions at 22\% (8\% for ARLoop candidate sequences, seed 8801). The ridge penalty is 0.025 times the mean diagonal of the centered input Gram matrix. ARLoop and LoopRes predict both the second- and third-loop writes for one-loop readout. Inkling and Dory predict only the third-loop interface.

For each path we backpropagate through the frozen readout and, when retained, the final loop. The loss combines cross-entropy on the correct option (weight 0.7), KL divergence from the three-loop option distribution to the shortcut distribution (weight 0.3), and an $L_2$ penalty on the trainable correction. The correction acts in a 48-dimensional input and 64-dimensional output subspace derived from the singular vectors of the first-loop states and the remaining fit errors. AdamW uses a learning rate of 0.009, weight decay of 0.04 and gradient clipping at 1. We merge the correction into the original weights and biases for inference. We fix the residual scales and correction gains from the 800-question fits using disjoint validation sets of 217 ARC-Challenge and 300 GSM8K training questions. Accuracy determines the choice, with option negative log-likelihood breaking ties. To match the number of optimizer steps in the 800-question fits, we cycle through the 72 fitting questions for 134 steps per path in ARLoop and 34 in the other models. We weight the mean squared correction penalty by 0.05. For Inkling's ARC-Challenge final-loop fit we apply this weight separately to the weights and the biases.

\paragraph{State-fitted predictors.}
On continuation and in the original four-option experiments, we fit the residual maps on up to 4,096 sampled token positions per checkpoint and task (seed 3671). Table~\ref{tab:four_models_ridge} presents the original four-option results. The GSM8K fits use the same 72 source questions as Table~\ref{tab:four_models}, with distractors and option order generated from a 264-question training draw. ARLoop and Dory use the unscaled ridge maps. For \inklingmini{} and LoopRes, we compared the original residual, its 0.65-scaled version and a correction at the answer position. The correction blends the original estimate with a ridge fit using both first-loop writes (weight 0.75 on the latter). Selection uses accuracy on 96 training-split validation questions, with agreement with three loops breaking ties. \inklingmini{} uses the scaled residual except on the GSM8K final-loop path, which uses the original residual. LoopRes uses the answer-position correction for direct ARC-Challenge readout and the scaled residual for its other paths. Table~\ref{tab:independent_options} reports an additional test of the state-fitted predictors.

\begin{table*}[t]
\centering
\caption{Four-option accuracy (\%) for the original state-fitted affine predictors. Each predictor uses 72 fitting questions and the state-based variants described above. The table uses the same checkpoints, 144 test questions and scoring rules as Table~\ref{tab:four_models}.}
\label{tab:four_models_ridge}
\small
\setlength{\tabcolsep}{4.6pt}
\begin{tabular}{@{}llrrrrr@{}}
\toprule
& & \multicolumn{2}{c}{One native loop} & \multicolumn{2}{c}{Two native loops} & \multicolumn{1}{c}{Three native loops} \\
\cmidrule(lr){3-4}\cmidrule(lr){5-6}\cmidrule(l){7-7}
Model & Task & Native & \phantom{Direct readout (ours)}\llap{\jump{1st loop only}} & Native & \phantom{Retain final loop (ours)}\llap{\jump{1st + last loops}} & Native \\
\midrule
ARLoop-0.6B & ARC-Challenge & 25.0 & \jump{29.9} & 30.6 & \jump{34.7} & 36.1 \\
 & GSM8K & 31.3 & \jump{36.1} & 30.6 & \jump{43.1} & 44.4 \\
\addlinespace[1.5pt]
LoopRes-1.7B & ARC-Challenge & 59.0 & \jump{66.7} & 75.0 & \jump{74.3} & 75.0 \\
 & GSM8K & 23.6 & \jump{28.5} & 31.9 & \jump{32.6} & 33.3 \\
\addlinespace[1.5pt]
\mbox{Inkling-26.4B} & ARC-Challenge & 92.4 & \jump{91.0} & 96.5 & \jump{95.8} & 97.2 \\
 & GSM8K & 47.2 & \jump{45.1} & 45.8 & \jump{51.4} & 48.6 \\
\addlinespace[1.5pt]
Dory-4B & ARC-Challenge & 80.6 & \jump{81.3} & 84.0 & \jump{84.0} & 84.0 \\
 & GSM8K & 41.0 & \jump{43.8} & 46.5 & \jump{47.9} & 48.6 \\
\bottomrule
\end{tabular}
\end{table*}

\paragraph{Readout interventions.}
The residual intervention and probe study uses answer-letter scores and unscaled ridge residuals on the same test questions, with 144 additional training questions for fitting probes and routers and 48 for calibration. The probes use random projections of the two answer-position writes with 96 coordinates per probe. The terminal-write interventions restore one native tensor while leaving the other predicted. Signed perturbations point toward or away from the first-loop write with amplitude 4\% of the joint-state RMS. We measure the margin between the correct and strongest incorrect option in the full model.

\begin{table*}[t]
\centering
\caption{Correct answers for the state-fitted predictors of Table~\ref{tab:four_models_ridge} on 384 additional questions per task. Within each row, all routes use the same questions, options and scoring rule.}
\label{tab:independent_options}
\small
\begin{tabular}{@{}llrrrrr@{}}
\toprule
Model & Task & Native one & \jump{1st loop only} & Native two & \jump{1st + last loops} & Native three \\
\midrule
ARLoop-0.6B & ARC-Challenge & 109 & \jump{92} & 124 & \jump{144} & 172 \\
 & GSM8K & 100 & \jump{127} & 121 & \jump{164} & 180 \\
LoopRes-1.7B & ARC-Challenge & 226 & \jump{281} & 294 & \jump{289} & 288 \\
 & GSM8K & 90 & \jump{142} & 160 & \jump{144} & 133 \\
\inklingmini{} & ARC-Challenge & 343 & \jump{346} & 357 & \jump{357} & 362 \\
 & GSM8K & 204 & \jump{206} & 214 & \jump{218} & 220 \\
\bottomrule
\end{tabular}
\end{table*}

On the original GSM8K generation task, ARLoop-0.6B scores 40.9\% with three native loops and five chain-of-thought demonstrations \citep[Tables 3 and 13]{yao2026arloop}. Dory1.0-4B Base scores 89.99\% with two loops and 89.84\% with four, using eight demonstrations and exact match on generated answers to all 1,319 test questions \citep[Table 1]{fishman2026dory}. Our Dory rows use the released post-trained BF16 checkpoint and a four-option readout.

\paragraph{Hardware and response timing.}
The four-checkpoint accuracy evaluations and continuation timings used NVIDIA L20X GPUs. Each timing comparison used one GPU at batch size one with key--value caching disabled. We report medians over seven prefixes after two warm-ups. In separate free-response GSM8K tests with \inklingmini{}, predictors are fitted on 96 prompts and 8,192 token positions. On 24 test questions, native two loops, retaining the final loop and native three loops take 283.0, 269.4 and 367.5 ms per response, respectively, including generation and stopping. On 256 test questions, retaining the final loop answers 106 correctly, compared with 94 for two native loops and 108 for three. The paired gain over two loops is 4.69 pp (95\% interval \([0.78,8.98]\)).

\noindent\begin{minipage}{\columnwidth}
\centering
\captionof{table}{Continuation forward time on the four released checkpoints (ms). \emph{1st + last loops} includes state prediction and two native loops.}
\label{tab:transfer_time}
\footnotesize
\setlength{\tabcolsep}{3.9pt}
\begin{tabular}{@{}lrrrr@{}}
\toprule
 & Native & \jump{1st + last loops} & \multicolumn{2}{c}{Full depth} \\
Model & 2 loops & ms & ms & Loops \\
\midrule
ARLoop-0.6B & 45.92 & \jump{46.33} & 69.39 & 3 \\
LoopRes-1.7B & 47.90 & \jump{47.91} & 66.12 & 3 \\
\inklingmini{} & 142.16 & \jump{144.26} & 198.84 & 3 \\
Dory-4B & 57.83 & \jump{58.05} & 77.88 & 3 \\
\bottomrule
\end{tabular}
\end{minipage}
\par\medskip

\begin{table*}[t]
\centering
\caption{Task accuracy and agreement with the native three-loop prediction (\%). Agreement is the fraction of examples on which a route makes the same prediction as the reference. Each row uses one checkpoint and a common set of 144 examples. Science and arithmetic are the ARC-Challenge and GSM8K runs from Table~\ref{tab:four_models}. Continuation uses the state-fitted predictors. The reference uses three native loops for each checkpoint.}
\label{tab:cross}
\small
\setlength{\tabcolsep}{4pt}
\begin{tabular}{@{}llrrrrrrr@{}}
\toprule
& & \multicolumn{3}{c}{Accuracy} & \multicolumn{4}{c}{Agreement with three native loops} \\
\cmidrule(lr){3-5}\cmidrule(l){6-9}
Model & Task & \jump{1st loop only} & \jump{1st + last loops} & Native 3 & Native 1 & \jump{1st loop only} & Native 2 & \jump{1st + last loops} \\
\midrule
ARLoop-0.6B & Continuation & 16.7 & 38.9 & 41.7 & 1.4 & 23.6 & 16.0 & 73.6 \\
 & Science & 27.1 & 33.3 & 36.1 & 40.3 & 46.5 & 47.2 & 70.8 \\
 & Arithmetic & 41.7 & 52.1 & 44.4 & 44.4 & 47.2 & 49.3 & 62.5 \\
LoopRes-Qwen3 & Continuation & 31.3 & 43.1 & 43.8 & 49.3 & 52.1 & 86.1 & 88.9 \\
 & Science & 70.1 & 77.8 & 75.0 & 63.2 & 72.2 & 87.5 & 84.0 \\
 & Arithmetic & 42.4 & 50.0 & 33.3 & 67.4 & 52.8 & 73.6 & 47.9 \\
\inklingmini{} & Continuation & 50.7 & 54.2 & 56.9 & 84.0 & 82.6 & 91.0 & 88.9 \\
 & Science & 91.7 & 97.2 & 97.2 & 93.8 & 92.4 & 99.3 & 100.0 \\
 & Arithmetic & 45.8 & 53.5 & 48.6 & 63.2 & 65.3 & 72.9 & 75.7 \\
Dory-4B & Continuation & 38.2 & 38.9 & 39.6 & 85.4 & 86.8 & 97.2 & 94.4 \\
 & Science & 79.2 & 84.0 & 84.0 & 92.4 & 88.2 & 97.9 & 97.2 \\
 & Arithmetic & 43.1 & 54.2 & 48.6 & 72.2 & 66.0 & 93.1 & 76.4 \\
\bottomrule
\end{tabular}
\end{table*}

\begin{table*}[t]
\centering
\caption{Readout diagnostics using state-fitted predictors on ARLoop, LoopRes and \inklingminiwithsize{}. The final write drives the largest change in agreement with the reference. Probes report accuracy before the first loop and after each native loop. The margin response is the mean absolute half-difference under equal positive and negative perturbations of 4\% joint RMS. The last column gives full-depth agreement after restoring only the last native write in \jump{1st loop only}.}
\label{tab:four_diagnostics}
\setlength{\tabcolsep}{5pt}
\begin{tabular}{@{}llrrr@{}}
\toprule
Model & Task & Linear probe accuracy (\%) & Margin response (earlier / last) & Last restored (\%) \\
\midrule
ARLoop-0.6B & ARC-Challenge & \(31.9\to25.0\to26.4\to27.1\) & 0.007 / 0.753 & 89.6 \\
 & GSM8K & \(25.7\to22.2\to27.8\to35.4\) & 0.004 / 0.536 & 90.3 \\
LoopRes-1.7B & ARC-Challenge & \(23.6\to34.0\to51.4\to54.9\) & 0.027 / 0.073 & 95.1 \\
 & GSM8K & \(31.2\to30.6\to34.7\to34.0\) & 0.029 / 0.039 & 89.6 \\
\inklingmini{} & ARC-Challenge & \(27.8\to88.2\to94.4\to95.8\) & 0.021 / 0.114 & 100.0 \\
 & GSM8K & \(23.6\to30.6\to38.9\to45.8\) & 0.018 / 0.093 & 91.0 \\
\bottomrule
\end{tabular}
\end{table*}

\begin{figure*}[t]
\centering
\includegraphics[width=\textwidth]{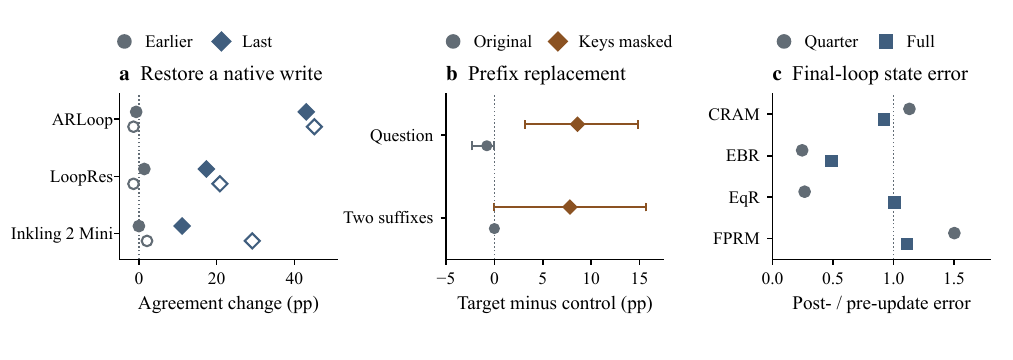}
\caption{State interventions. (a) Change in full-depth agreement after restoring one native write. Filled markers denote ARC-Challenge and open markers denote GSM8K (144 items per setting). (b) Counterfactual hit-rate difference between targeted and unrelated donors, by question or by two-suffix family. Intervals use a family bootstrap. Masking is exploratory. (c) For CRAM, EBR, EqR and FPRM on the same 192 Sudoku-Extreme puzzles, we report the median ratio of post- to pre-update RMS state error after adding one quarter or all of the prediction error to the native input.}
\label{fig:diagnostics}
\end{figure*}

\paragraph{Residuals and caching.}
In the independent residual protocol we restore one of two late writes while leaving the rest predicted. Figure~\ref{fig:diagnostics}a gives both restoration effects over six model--task pairs. The last write also changes the correct-versus-incorrect-option margin more than the earlier one in all six equal-amplitude perturbation comparisons. A timing intervention in ARLoop replaces the same history with a first-loop copy before the final loop, or after its first nine layers. Agreement with the reference is 39.34\% and 93.21\%, respectively.

\paragraph{Prefix cache depth in Dory.}
In a separate Dory-4B ablation we vary current-token depth and the depth at which prefix keys and values are computed. Each setting uses the same 144 examples per task. With one current-token loop, using the second-loop cache increases agreement with the two-loop cached reference from 86.1 to 91.0\% on continuation, from 92.4 to 100.0\% on ARC-Challenge, and from 67.4 to 85.4\% on GSM8K. The paired 95\% bootstrap intervals for the changes are \([-0.7,10.4]\), \([3.5,11.8]\), and \([11.1,25.0]\) pp, respectively, from 5,000 resamples. On ARC-Challenge, one loop with the deep cache gives 84.72\% accuracy and matches the two-loop reference on every answer. Two loops with the shallow cache give 79.86\% accuracy. The cache is computed in advance, and this intervention counts only the loops run on the current token.

\paragraph{Long-horizon structured reasoning.}
The puzzle experiments use frozen public weights. TRM uses the \href{https://huggingface.co/MaximeILL/GAR_ckpts/tree/26c3f66aa6554991e8f35f1950b54cd7d84874d0}{TRM-MLP Sudoku checkpoint} at EMA step 65,100 and the \href{https://huggingface.co/ThomasHeim/HRM-Reproduction-Checkpoints/tree/c20fbd0d7f52a69e3343a7f6dff66e60bca34524}{reproduced TRM-Attention Maze checkpoint} at EMA step 156,240. CHARM uses the \href{https://github.com/water-vapor/charm}{authors' ARC-AGI-2 configuration} and released EMA weights at step 2.38M. Appendix~\ref{app:energy} reports separate fixed-point probes on Sudoku-Extreme. Only the late-state predictors are fitted for these comparisons.

Figure~\ref{fig:solvers} uses 16 native updates as the reference for each model. We reuse the fitted CHARM predictor and its 640 fitting tasks. For TRM we fit the predictors for 350 updates on 768 examples using state regression and cross-entropy with the native output distribution. EBR uses the released Maze-Hard EMA weights \citep{chen2026ebr} and one trajectory per maze. A shared three-block attention predictor with two output heads learns RMS-normalized residual updates from 640 training mazes for 1,500 optimizer steps, using state regression, teacher-output cross-entropy and supervised cross-entropy. All EBR routes use paired initial seeds. Each loop runs one native descent segment of 21 steps, which may include noise and projection. The full EBR reference runs 16 loops, and both shortcuts retain the first loop. Times include embedding, prediction, native updates and readout. For CHARM, EBR and TRM we report median synchronized forward time per input at batch sizes 32, 16 and 8, respectively.

For EBR, the difference between the two-loop route and full depth is \(-0.30\) pp with a paired 95\% bootstrap interval of \([-0.90, 0.30]\) pp. On the 744 test mazes outside the initial screen, it is \(-0.40\) pp with an interval of \([-1.08, 0.27]\) pp. The two-loop route scores 0.30 pp above the native two-loop prefix, with a paired interval of \([-0.40, 1.00]\) pp.

\begin{table*}[t]
\centering
\small
\caption{Matched one- and two-loop controls for CHARM, TRM and EBR in Figure~\ref{fig:solvers}. Entries are score (\%) / inference time (ms) per augmented grid for CHARM and per maze for TRM and EBR. EBR timing includes the energy-gradient evaluations.}
\label{tab:solver_routes}
\setlength{\tabcolsep}{3.6pt}
\begin{tabular}{@{}lrrrrrr@{}}
\toprule
& \multicolumn{2}{c}{One loop} & \multicolumn{2}{c}{Two loops} & \multicolumn{2}{c}{Full depth} \\
\cmidrule(lr){2-3}\cmidrule(lr){4-5}\cmidrule(l){6-7}
Model & Native & \jump{1st loop only} & Native & \jump{1st + last loops} & Native & Loops \\
\midrule
CHARM & 24.72 / 4.45 & \jump{26.39 / 4.47} & 35.14 / 8.88 & \jump{35.14 / 8.89} & 35.97 / 70.83 & 16 \\
TRM & 79.10 / 2.00 & \jump{78.30 / 2.02} & 86.40 / 3.95 & \jump{86.10 / 3.97} & 86.60 / 31.32 & 16 \\
EBR & 85.00 / 6.73 & \jump{83.60 / 6.85} & 86.00 / 13.39 & \jump{86.30 / 13.58} & 86.60 / 109.88 & 16 \\
\bottomrule
\end{tabular}
\end{table*}

\paragraph{First and final loops on Sudoku.}
We use the \href{https://huggingface.co/pengxiang/InfiLoop-Sudoku/tree/4b6e8ad1f4f92e49290e73c514ff7329f79cc6b2}{released InfiLoop checkpoint} at EMA step 71,610 \citep{li2026infiloop}. One loop is one model segment. We use the authors' fixed-depth setting of six high-level cycles and twelve low-level updates per cycle, or 78 calls to the shared two-layer block per loop. All routes use this internal depth, including our 16-loop reference.

On 384 training puzzles we fit a position-shared ridge map from the high state, low state and input embedding after the first loop to the changes in both states through loop 15. Among three configurations, InfiLoop at the six-by-twelve setting had the highest two-loop accuracy on 232 validation puzzles, with a residual scale of \(1/8\) for both states. We retain the first loop's memory weight and multiply it by the predicted high state to obtain the memory numerator. Both shortcuts start from this prediction. Direct readout applies the frozen output head, and the two-loop route runs the final native loop before readout.

\begin{table}[t]
\centering
\small
\caption{InfiLoop on 2,048 Sudoku-Extreme puzzles (Figure~\ref{fig:solvers}b). Time is the median per grid on one NVIDIA L20X at batch size 16. Routes were interleaved for 24 synchronized timings after eight warm-ups, including embedding, state prediction where used, and readout.}
\label{tab:sudoku_routes}
\setlength{\tabcolsep}{3pt}
\begin{tabular}{@{}lrrr@{}}
\toprule
Route & Exact (\%) & ms/grid & Speedup \\
\midrule
Native one loop & 52.73 & 5.20 & \(16.08\times\) \\
1st loop only & 52.73 & 5.21 & \(16.04\times\) \\
Native two loops & 72.27 & 10.40 & \(8.03\times\) \\
1st + last loops & 71.53 & 10.50 & \(7.96\times\) \\
Native 16 loops & 92.29 & 83.58 & \(1.00\times\) \\
\bottomrule
\end{tabular}
\end{table}

In the published EBR sweep, gradient-based stopping preserves 99.92\% pass@8 while reducing computation from 6.60 to 0.61 TFLOPs per Sudoku-Extreme puzzle. These values use eight samples per puzzle and the minimum-energy answer, and include the gradient checks \citep[Table C.2]{chen2026ebr}.

Using the TRM Sudoku checkpoint \citep{jolicoeur2025trm}, we obtain 47.80\% accuracy by predicting loop 7 from loop 1 and retaining loop 8, versus 54.25\% for two native loops and 81.40\% for eight. Predicting only the working state raises the shortcut to 54.90\%. On a separate 800-problem Maze-Hard test, the two-loop native and direct-readout TRM paths both score 86.25\%, while 16 loops score 86.50\%. In that protocol FPRM direct readout from a prediction based on loop 16 scores 84.88\%, matching the 25-loop reference in fixed-path accuracy while changing two correct and two incorrect predictions. In an exploratory 16-view ARC-AGI-2 sweep with CHARM \citep{zhao2026charm}, we obtain 38.89\% pass@2 by keeping the first eight and last loops and predicting the intervening state, compared with 36.81\% for the first nine native loops.

\section{Latent Continuation Protocol}
\label{app:cot}

For \inklingmini{}, we use constructed relation chains whose factual lookup length is set by the question. For instance, a person is mapped to a key, that key to a room, and the room to an object. The native depth sweep contains 72 families and 144 questions per hop count. Answers are scored over the full vocabulary. Three-hop accuracies for depths 1, 2, 3 and 5 are \(97,116,139,142\) out of 144. The three-versus-two gain is 15.97 pp with a family-paired interval \([10.42,21.53]\). One-loop performance is already \(137/144\) for two-hop questions.

For fixed-state continuation we use 64 held-out families with two suffixes each and score five answer options. The separately fitted head trained to predict the final state gives \(79/128\). Directly reading out the fixed predicted penultimate state gives \(77/128\), and the same state followed by the real final loop gives \(115/128\). The latter corrects 38 of the 51 direct-readout errors in that intervention. The shortcut's paired improvement over native two loops is 7.81 pp with family bootstrap interval \([1.56,14.06]\). The suffixes have different downstream mappings and are processed individually in the fixed-state comparison.

For prefix interventions, the donor sees only a prefix which states a relation about a key. The targeted donor changes that key, and the control donor changes a different person's fact. In the recipient we replace the prefix state at the final-loop interface, while preserving the recipient's wording and suffix writes within each contrast. The suffix specifies the remaining relations. Predictor fitting, subspace fitting, and testing use disjoint 64-family sets. After inspecting the original results, we test a recipient prefix in which all stated key colors are replaced by gray. On these questions the per-question effect of targeted predicted-prefix replacement is 8.59 pp with family-paired interval \([3.13,14.84]\). Both suffixes switch in 7 targeted and 2 control families. Explicitly supplying the edited key in text switches 127 of 128 masked questions and both suffixes in 63 of 64 families. Replacing a norm-matched learned four-dimensional subspace at the marker changes at most one more prediction than its random control in the tested main contrast. This separates whole-prefix intervention from a claim about an isolated semantic variable.

\section{Dynamical Analysis and Output Perturbations}
\label{app:theory}

\paragraph{Fast--slow error.}
Subtract the moving affine equilibrium from \(v_{k+1}\) to obtain
\[
 e_{k+1}=Ae_k+v_*(u_k)-v_*(u_{k+1})+r_k.
\]
The triangle inequality gives the error bound for one loop in Section~\ref{sec:theory}. A geometric sum produces Equation~\eqref{eq:fast}. The bound for reconstruction then follows from Lipschitz continuity. In a differentiable feedback module \(v^+=\Lambda v+\phi(u,v)\), where \(\Lambda\) is a linear skip and \(\phi\) the remaining update, a local candidate bound for the fixed-\(u\) gain is \(\|\Lambda\|+\operatorname{Lip}_v(\phi)\). The state in this bound includes any messages passed between loops.

\paragraph{Reference margin and local divergence.}
Integrating each absolutely continuous margin and applying the dual-norm inequality yields Equation~\eqref{eq:margin}. For \(G\) twice continuously differentiable near \(s\), the small-error divergence is
\[
 \begin{aligned}
 \KL(p\Vert \operatorname{softmax}G(s+\xi\delta))
 &=\frac{\xi^2}{2}(J\delta)^\top C_p(J\delta)\\
 &\quad+O(\xi^3),
 \end{aligned}
\]
where \(p=\operatorname{softmax}G(s)\), \(C_p=\operatorname{diag}(p)-pp^\top\), and \(J\) is the Jacobian at \(s\). Set \(\Delta=G(s+\delta)-G(s)\) and \(p_\tau=\operatorname{softmax}(G(s)+\tau\Delta)\) for \(\tau\in[0,1]\). The finite-amplitude divergence is exactly the integral of \((1-\tau)\Delta^\top C_{p_\tau}\Delta\) over \([0,1]\), by the integral expansion of log-sum-exp. At one-eighth error scale, the rank correlation of the quadratic estimate with measured KL ranges from 0.612 to 0.882 across the six settings for ARLoop, LoopRes and \inklingmini{}.

\paragraph{Energy and stability.}
At a smooth energy minimum, linearizing the gradient update gives \(I-\alpha\nabla^2 E_x\). The linear update contracts in each direction with \(0<\alpha\lambda_i<2\). If a smooth fast equilibrium satisfies \(\partial_v E(u,v_*(u))=0\) with positive definite block \(H_{vv}\), implicit differentiation gives \(dv_*/du=-H_{vv}^{-1}H_{vu}\). This gives a local affine approximation to the fast equilibrium as the slow coordinate shifts.

An autonomous map \(F_x\) with a fixed point \(s_*\) and stable Jacobian \(A_*=DF_x(s_*)\), meaning spectral radius below one, has a quadratic local Lyapunov function. For positive definite \(Q\), the series \(M=\sum_{i\ge 0}(A_*^i)^\top Q A_*^i\) solves \(M-A_*^\top M A_*=Q\). The function with value \(\tfrac12\delta^\top M\delta\) for error \(\delta\) measures local stability. A fixed-step Euclidean energy representation additionally requires a symmetric residual Jacobian in the coordinates used. In EqR and FPRM, the observed two-direction finite-difference antisymmetry has a consistent nonzero sign across three scales for 60/64 and 63/64 tested inputs, respectively. These are local numerical tests in the native state coordinates.

\section{Energy and Fixed-Point Experiments}
\label{app:energy}

\paragraph{Four-loop language model.}
The model has 5,623,808 parameters and 128 coordinates each for the slow and fast state. The slow state receives a residual update bounded by 0.15 per coordinate. Two 4,080-step runs use the same seed, data order, batch of 256 windows and 133,693,440 nominal positions. The control copies states and the learned branch predicts the late slow state. Both branches share the form \(v^+=0.25v+0.75g_\theta(a_x,u^+)\), where \(u^+\) is the updated slow state. At fixed \(u^+\) this update gives an energy ratio of \(1/16\). The logged energy is averaged over fast coordinates, which leaves the ratio unchanged. At a fixed predicted slow state, reconstruction improves agreement from 81.57 to 83.01\% for direct readout and from 89.18 to 89.61\% with the final loop. These tests use 131,072 WikiText-2 target positions.

\paragraph{Predicting EqR's working state.}
We retain the first-loop answer state \(z_H\) and predict a residual for the working state \(z_L\) before one native final loop. The direct readout depends only on \(z_H\) and matches the first-loop prediction on every cell. Across two held-out sets totaling 6,144 Sudoku-Extreme puzzles, applying the final loop to the predicted working state solves 2,279 grids (37.09\%), compared with 2,276 (37.04\%) for two native loops.

\paragraph{Same-puzzle solver probes.}
CRAM, EBR, EqR, and FPRM use the same 192 training and 192 held-out Sudoku-Extreme puzzles. We fit one affine map shared over positions and replay the native eighth update with the original randomness where applicable. An update is a sweep for zero-temperature CRAM, a segment of 21 projected noisy gradient steps for EBR, an outer loop for EqR, and a damped update for FPRM. EBR and EqR show decreases of 17.71 and 13.54 pp for the shortcut that retains the final loop versus two native updates (Table~\ref{tab:energy}). The respective paired 95\% intervals for shortcut minus native are \([-23.44,-12.50]\) and \([-19.27,-7.81]\). Figure~\ref{fig:diagnostics}c shows that a perturbed state can contract even where the answer does not recover.

The puzzle backbones are frozen. We use \href{https://github.com/Prefrontal-AI/CRAM}{CRAM's released Sudoku-Extreme seed-zero weights}, the \href{https://huggingface.co/wxingxu/ebr/tree/da0172a51409281079094bb5f6d19b5bdc1c44bb}{EBR Sudoku release at revision \texttt{da0172a}}, the \href{https://huggingface.co/locuslab/EqR-model}{EqR Sudoku weights}, and the \href{https://huggingface.co/fixed-point-reasoners/fprm}{FPRM EMA weights at step 78,120}. The SHA-256 prefixes of the loaded weights for this split are \texttt{8ec939cd}, \texttt{541e5748}, \texttt{258b1f0a}, and \texttt{6daec5f4}, respectively.

\noindent\begin{minipage}{\columnwidth}
\centering
\captionof{table}{Sudoku-Extreme results on 192 puzzles. Each cell shows exact / blank-cell accuracy (\%). CRAM uses zero-temperature sweeps here.}
\label{tab:energy}
\footnotesize
\setlength{\tabcolsep}{3pt}
\begin{tabular}{@{}lrrrr@{}}
\toprule
Path & CRAM & EBR & EqR & FPRM \\
\midrule
Native 1 & 0.0 / 24.4 & 50.5 / 74.4 & 5.2 / 60.1 & 0.0 / 17.0 \\
1st loop only & 0.0 / 24.4 & 50.5 / 73.9 & 4.7 / 59.9 & 0.0 / 18.8 \\
Native 2 & 0.0 / 25.6 & 68.8 / 83.6 & 34.4 / 72.1 & 0.0 / 44.6 \\
1st + last loops & 0.0 / 25.6 & 51.0 / 73.3 & 20.8 / 67.4 & 0.0 / 25.4 \\
Copy + final & 0.0 / 25.6 & 69.3 / 84.7 & 33.3 / 71.8 & 0.0 / 44.6 \\
Native 8 & 0.0 / 26.7 & 96.9 / 98.3 & 65.1 / 82.5 & 16.7 / 68.6 \\
\bottomrule
\end{tabular}
\end{minipage}
\par\medskip

\paragraph{Full CRAM search.}
With the author's annealing schedules, all paths solve the 192 puzzles. In this setting the original and learned initializations take 45.13 and 77.86 s, respectively. An independent experiment with 2,048 puzzles keeps accuracy at 100\% and reduces wall time from 113.77 to 107.14 s, with estimated operations increasing by 4.33\%. Each comparison uses the same solver and schedule across its initializations.

\end{document}